\documentclass[10pt]{article}
\usepackage[preprint]{tmlr}

\usepackage{graphicx}
\usepackage{amsmath,amssymb}
\usepackage{booktabs}
\usepackage{longtable}
\usepackage{enumitem}
\usepackage{xcolor}
\usepackage{hyperref}
\hypersetup{hidelinks}   
\usepackage{url}
\usepackage[capitalize]{cleveref}
\usepackage{tikz}
\usetikzlibrary{arrows.meta,positioning,calc,fit,backgrounds}
\definecolor{csblue}{RGB}{31,78,121}
\definecolor{csred}{RGB}{178,34,34}
\title{CurveShift: Is Agent Progress Scalar? Separating Level from Shape}

\author{%
Hanwen Xing$^{1}$ \quad Pengyun Wang$^{2}$ \quad BingXu Meng$^{3}$ \quad Kumail Alhamoud$^{4}$\\
Xiang Li$^{5}$ \quad Jicheng Wang$^{6}$ \quad Xin Yu$^{7}$ \quad Xinyang Han$^{3}$\\
Xiaomin Li$^{8}$ \quad Philip Torr$^{9}$ \quad Yuexing Hao$^{4}$
\\[0.4em]
\normalfont\small\itshape
$^{1}$University of Southern California \quad $^{2}$University of Chicago \quad $^{3}$University of California, Berkeley\\
$^{4}$Massachusetts Institute of Technology \quad $^{5}$Stanford University \quad $^{6}$University of California, Davis\\
$^{7}$Pennsylvania State University \quad $^{8}$Harvard University \quad $^{9}$University of Oxford}

\begin{document}
\maketitle

\begin{abstract}
Progress in large language models is often summarized using a single scalar measure, such as a time horizon, a latent ability estimate, or an aggregate benchmark score. These summaries capture the overall
performance, but they do not test whether progress is distributed differently across task difficulty. 
We find that most of the apparent shift in gains toward harder tasks does not reflect a change in the
shape of the difficulty-response curve. On METR time-horizon data, a single Rasch model with rising
ability reproduces this pattern, so it is largely explained by ceiling effects rather than a
qualitative change in capability. This echoes how the choice of metric can make claimed emergent
abilities look like a property of the models themselves. We then identify a smaller hard-task effect
that survives this control. Isolating it is difficult on agentic benchmarks, because newer models are
usually run with newer agentic harnesses, so a gain on hard tasks cannot be assigned to the model or
its scaffold. We break the confound with LiveCodeBench, a public competitive programming benchmark
that runs no agentic scaffold while pairing dated models with an exogenous difficulty ordering. After
accounting for the rise in overall ability, models released after September 2024 still gain on the
hardest problems beyond what their easy and medium performance predicts, by about $+0.40$ logits
under our most conservative assumption, raising the hard-problem solve rate from roughly $18\%$ to
$25\%$. The effect is led by the strongest reasoning models and holds for hard tasks that need only
short reasoning, not autonomy over long horizons. We present this as a result specific to competitive
programming, since our clean identification rests on a single coding benchmark. We release the LiveCodeBench Difficulty Panel ($66$ dated models $\times$ 1{,}055 problems) and our analysis code.\footnote{\url{https://github.com/harvenstar/CurveShift}}

\textbf{Keywords:} item response theory, differential item functioning, language model evaluation, benchmark saturation, emergent abilities, capability forecasting, competitive programming, code generation, reasoning models, test-time compute, construct validity, scaling
\end{abstract}

\section{Introduction}
\label{sec:intro}

\begin{figure}[!t]
\centering
\includegraphics[width=\linewidth]{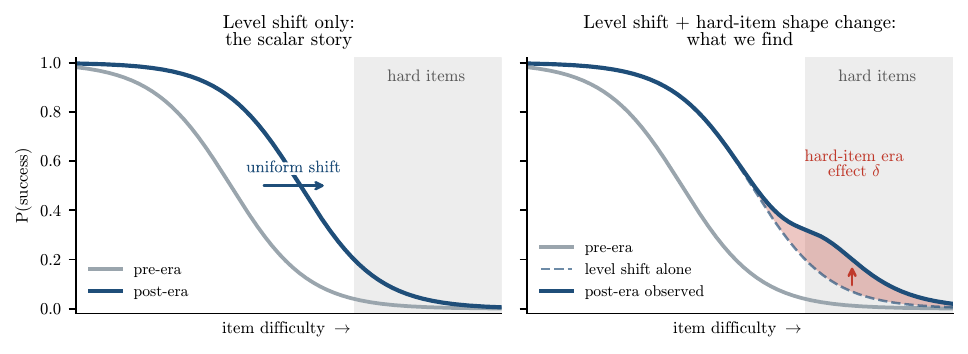}
\vspace{-6pt}
\caption{Level-shape conflation, schematically. Left: a uniform upward shift in ability moves the
whole difficulty-response curve; this is the scalar story. Right: the same level shift plus a gain
concentrated on hard items. Both raise the average score, so a scalar summary cannot tell them apart. The shaded gap is the hard-item era effect
$\delta$ that our design isolates.}
\label{fig:schematic}
\vspace{-8pt}
\end{figure}

Progress in large language model and agent capabilities is often summarized by a single number.
Leaderboards report average solve rates; the Model Evaluation and Threat Research (METR) time-horizon metric summarizes longitudinal
improvement using a scalar doubling time~\citep{metr2025horizon}; latent ability models fit a single
rising curve to benchmark performance over model generations~\citep{epoch_rosetta}.
These scalar summaries are useful, but they conflate two distinct phenomena: a uniform upward shift
in ability and a disproportionate gain at high difficulty both raise the average score
(\cref{fig:schematic}), but without separating these effects, we cannot tell which has occurred, or whether both have. 
We introduce \emph{CurveShift}, an analysis that separates an overall increase in ability from a change in the difficulty-response curve concentrated on hard items. Our main finding is
that, after separating overall ability growth from difficulty-specific gains, most of the reported shift toward harder tasks is not a shape
change at all but an artifact of a single rising ability, and only a small residual effect remains.

If models released after September 2024 improve more on hard tasks than on easy ones, this indicates a change in the difficulty-response profile of deployed systems. In particular, they solve hard problems more often than their easy and medium performance would predict. 
Conversely, if gains are proportional across difficulty, the capability frontier is moving
uniformly and scalar summaries are adequate.

Item response theory (IRT) provides the statistical language for this question~\citep{lord1980irt}.
The Rasch model~\citep{rasch1960} treats performance as a function of a single latent ability and a
fixed item difficulty; a 2-parameter logistic (2PL) model~\citep{birnbaum1968} additionally allows
item discrimination to vary.
Differential item functioning (DIF) tests whether the relationship between ability and item
difficulty changes across groups or eras~\citep{holland1993dif,swaminathan1990dif}, which is precisely the test we
need. 
We use DIF in the broad sense of era-dependent item functioning: our parameter is a group-by-difficulty shift concentrated on hard items that is constant across ability levels, not a group-by-ability slope interaction.

Applying IRT longitudinally to agentic benchmarks introduces an identification problem.
Models released after September 2024 are almost never evaluated with harnesses developed before that period. 
As~\citet{hal2025} document, the choice of agentic harness can affect scores as much as the choice of model.
Newer models are usually evaluated with newer harnesses, so the effects of the model and the harness are nearly collinear and difficult to separate. 
Any DIF at the era level estimated from METR or SWE-bench trajectories therefore conflates a model
capability improvement with a harness improvement, and the two cannot be separated using that
data.
We avoid this confound by using LiveCodeBench~\citep{livecodebench}. 
In LiveCodeBench, 
models produce code directly from a problem statement, in the tradition of direct program-synthesis
evaluations~\citep{austin2021programsynthesis,chen2021codex}, without an agent harness, tool-calling loop,
or a multistep scaffolding layer. 
The human contest origin of its problems provides an exogenous difficulty ordering that is
independent of any model.
This makes LiveCodeBench the identifying benchmark for our analysis. It isolates the effect at the model-native layer, where models generate directly with no harness, and by construction leaves attribution on agentic benchmarks, where model and harness move together, unresolved.

Our clean identification holds for hard tasks that need only short reasoning paths, as measured on
LiveCodeBench, and not for autonomous agent tasks over long horizons. 
Because the analysis uses a coding benchmark, our result is specific to competitive programming. 
The reasoning era in our framing bundles several concurrent changes, so we do not isolate reasoning
training as a causal factor.
Our estimand is the performance of deployed models as used in practice, including any native
inference-time reasoning they perform.

The argument runs in three steps: we deflate the apparent trend against a scalar null, localize the residual to a specific evaluation layer, then interpret how far the evidence licenses a conclusion.
We make three contributions. 
First, our primary result challenges the common interpretation of the apparent shift in gains toward harder tasks. Although this pattern has been taken as evidence that the capability frontier is changing shape, we show that a Rasch model with a single rising ability parameter largely reproduces it. The pattern is therefore explained mostly by ceiling and discrimination effects, rather than by a change in the difficulty-response curve. 
Second, we identify a smaller signal that remains after controlling for the overall rise in ability. Models released after September 2024 show differential item functioning on the hardest items, meaning that they perform better than their easy and medium results would predict. Because a free two-parameter logistic model is not stably identified on these data, we estimate the effect using anchored sensitivity analysis. The hard-item effect remains positive even under the conservative assumption that hard items are no more discriminating than easy and medium ones.
Third, we argue that this question can only be settled by the right kind of benchmark. Model era
and harness era are collinear in standard agentic evaluation data, which leaves a shape change
concentrated at specific difficulties irresolvable there, while a no-harness benchmark with an exogenous
difficulty ordering separates model capability from harness improvement.

\section{Related Work}
\label{sec:related}

\paragraph{Scalar summaries of agent capability progress.}
The dominant mode of reporting capability progress is a single aggregate trend, often over large benchmark
suites~\citep{srivastava2022bigbench,liang2022helm}. \citet{metr2025horizon} define a time-horizon metric and report a doubling time that has been
accelerating, but the analysis summarizes capability as a single scalar and does not ask whether
gains are uniformly distributed across task difficulty.
\citet{epoch_rosetta} use IRT to stitch performance across heterogeneous benchmarks into a unified
latent ability curve, detect acceleration, and attribute it to post-era models concentrated in
verifiable domains; however, they extract a single latent ability and do not test for
a shape change concentrated at specific difficulties or DIF at the era level.
Our work complements both: the same aggregate trend is consistent with a pure level shift and with
disproportionate gains on hard tasks, and distinguishing the two requires the DIF analysis we provide.

\paragraph{Measurement artifacts in capability claims.}
A recurring lesson in LLM evaluation is that an apparent qualitative change can be an artifact of the
measurement. Emergent ability claims were originally framed as abrupt scale-dependent capabilities~\citep{wei2022emergent}; \citet{schaeffer2023mirage} show that emergent abilities, the sharp jumps in capability
with scale, are largely an artifact of nonlinear or discontinuous metrics rather than a property of
the models or their scaling laws. Our deflation result applies the same critique across a different axis: the apparent
migration of gains toward harder tasks is largely reproduced by a scalar null, a ceiling and
discrimination artifact, not a change in the difficulty-response curve. We differ in where we stop.
The critique of emergence ends at deflation, while we go on to isolate a residual shape change that
survives the scalar control. Our contribution is not only that the naive reading is an artifact, but
that a smaller real effect remains underneath it. Framed in measurement terms, this is a question of
construct validity~\citep{bean2025constructvalidity} and benchmark choice~\citep{dehghani2021benchmarklottery}: whether a scalar score validly represents the shape of capability and not only
its level.

\paragraph{Psychometrics and exogenous difficulty anchoring.}
Item response theory has a track record in NLP evaluation. \citet{lalor2016irt} use it to build
evaluation scales, and \citet{rodriguez2021leaderboards} apply it to NLP leaderboards, jointly
modeling item difficulty and discrimination to find the examples that best separate systems;
\citet{tinybenchmarks} and the anchor point method of \citet{vivek2024anchor} use IRT to compress
benchmarks at a single snapshot.
A central challenge in these applications is that item difficulties are estimated from model
responses, which is circular when the goal is to measure how capability relates to difficulty.
\citet{bridge2026} estimate latent difficulty from model responses and then calibrate that scale
against human completion time, so their difficulty ordering is model-derived rather than
model-independent. Their finding that latent difficulty is linear in the logarithm of human
completion time supports the use of that quantity as a difficulty scale, and our METR analysis uses
it that way. We avoid the circularity by taking difficulty labels that humans assigned before any
model was evaluated.
\citet{agentpsychometrics} decompose agentic coding benchmark performance into separate LLM and
scaffold ability components using IRT at a single time point.
Our analysis is longitudinal and introduces a different decomposition: we separate the scalar
ability trend from DIF that varies by era, a test of measurement invariance across eras~\citep{vandenberg2000measurement}, applying the exogenously anchored difficulty scale to ask
whether the shape of the difficulty-response curve changes across the eras before and after
reasoning.

\paragraph{Scaffold confounding and identification.}
\citet{hal2025} show empirically that scaffold choice can move scores by as much as model choice.
When a new generation of models arrives alongside new scaffolding strategies, observed improvements
reflect both factors simultaneously, a general concern in LLM-agent evaluation~\citep{liu2023agentbench,zhou2023webarena,xie2024osworld}.
We formalize this as an identification problem: where model era and scaffold era are collinear,
a shape change concentrated at specific difficulties cannot be identified without additional assumptions, and moving to
a no-scaffold benchmark eliminates the confound by construction.
\citet{ockbench} provide a related motivation: scalar accuracy can conceal substantial variation in
token efficiency across models with identical solve rates, illustrating that scalar summaries hide
structure that matters for downstream conclusions.

\section{Data and Method}
\label{sec:data-and-method}

\begin{figure}[t]
\centering
\begin{tikzpicture}[
    font=\footnotesize,
    >={Stealth[length=2mm]},
    box/.style={rounded corners=2pt, draw=black!35, fill=white, align=center,
                text width=28mm, minimum height=9mm, inner sep=3pt},
    reject/.style={box, draw=black!22, fill=black!4, text=black!50},
    ident/.style={box, draw=csblue, fill=csblue!8, line width=0.6pt},
    method/.style={box, draw=csblue!70, fill=csblue!5},
    robust/.style={box, draw=black!40, fill=white},
    result/.style={box, draw=csred, fill=csred!7, line width=0.7pt, text width=36mm},
    panel/.style={rounded corners=4pt, draw=black!12, fill=black!2, inner sep=6pt},
    ptitle/.style={font=\scriptsize\bfseries, text=black!75},
    lbl/.style={font=\scriptsize, text=black!50},
    ar/.style={->, draw=black!45, line width=0.55pt},
  ]

  \node[reject] (metr) at (0,0)
        {METR time horizon\\[1pt]{\scriptsize\color{black!45}$\times$ scaffold confound}};
  \node[reject, below=3mm of metr] (swe)
        {SWE-bench Verified\\[1pt]{\scriptsize\color{black!45}$\times$ scaffold confound}};
  \node[ident, below=6mm of swe] (lcb)
        {\textbf{LiveCodeBench}\\[1pt]{\scriptsize\textcolor{csblue}{\checkmark} no scaffold,\\exogenous difficulty}};

  \node[method, right=32mm of metr] (level)
        {\textbf{Level}\\[1pt]{\scriptsize freeze ability on\\easy + medium items}};
  \node[method, below=8mm of level] (shape)
        {\textbf{Shape}\\[1pt]{\scriptsize hard-item era effect $\delta$,\\pinned $\alpha_{\mathrm{hard}}$}};

  \node[robust, right=26mm of level] (rob)
        {$\alpha_{\mathrm{hard}}$ grid\\[1pt]+ cluster bootstrap};
  \node[result, below=8mm of rob] (res)
        {most of the trend is a \textbf{level} artifact;\\[2pt]
         residual \textbf{shape} gain $\delta \approx +0.40$ logits on hard items};

  \begin{scope}[on background layer]
    \node[panel, fit=(metr)(swe)(lcb)] (p1) {};
    \node[panel, fit=(level)(shape)] (p2) {};
    \node[panel, fit=(rob)(res)] (p3) {};
  \end{scope}
  \node[ptitle, above=0.5mm of p1.north] {Identification};
  \node[ptitle, above=0.5mm of p2.north] {CurveShift decomposition};
  \node[ptitle, above=0.5mm of p3.north] {Estimation and result};

  \draw[ar, draw=csblue, line width=0.8pt]
        (lcb.east) -- node[lbl, text=csblue, below=0.3mm] {data} ($(lcb.east)+(13mm,0)$)
        |- (p2.west);
  \draw[ar] (level) -- (shape) node[midway, right=0.5mm, lbl] {freeze $\theta$};
  \draw[ar] (shape.east) -- ($(shape.east)+(7mm,0)$) |- (rob.west);
  \draw[ar] (rob) -- (res);

\end{tikzpicture}
\vspace{-6pt}
\caption{Study design. On agentic benchmarks (METR, SWE-bench) model era and scaffold era move
together, so a hard-item gain is not identified. LiveCodeBench runs no scaffold and carries an
exogenous human difficulty ordering, which identifies the effect. From this single panel, CurveShift
freezes ability on easy and medium items (the level) and, holding that ability fixed, estimates a
hard-item era effect $\delta$ with the discrimination $\alpha_{\mathrm{hard}}$ pinned rather than fit
(the shape). We report $\delta$ over a grid of
$\alpha_{\mathrm{hard}}$ with cluster bootstrap intervals. Most of the aggregate trend is a level
artifact; a residual shape gain of about $+0.40$ logits on hard items survives.}
\label{fig:design}
\vspace{-8pt}
\end{figure}
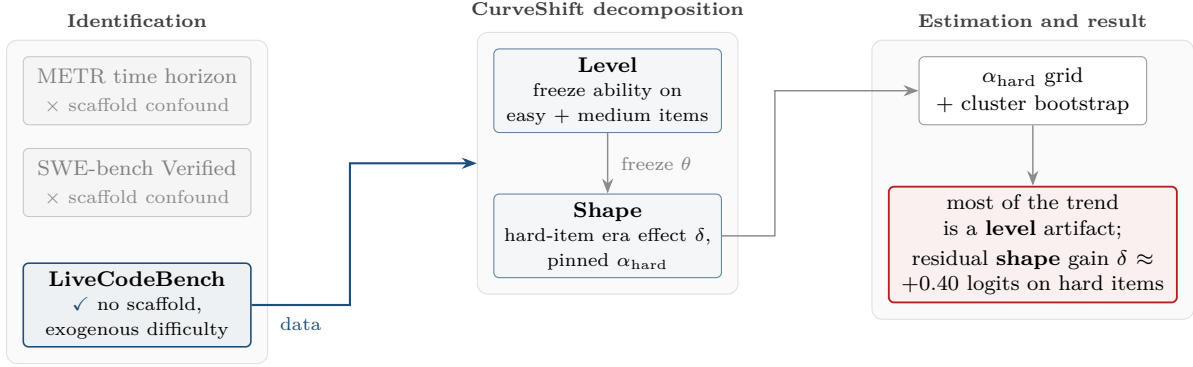

\Cref{fig:design} summarizes the design. We identify the effect on a no-scaffold benchmark, use
CurveShift to separate a level shift in ability from a hard-item shape change, and report the shape
effect over a grid of pinned discrimination with cluster bootstrap intervals.

\subsection{Data}
\label{subsec:data}

We assemble three public result matrices, each providing per-model per-item binary outcomes with
difficulty labels assigned by humans independently of the models under evaluation. This exogenous
labeling avoids the circularity that arises when difficulty is inferred from the same model
population whose progress is being measured. For METR the label is human completion time.
\citet{bridge2026} report that latent item difficulty is linear in its logarithm, which supports
its use as a difficulty scale.

\paragraph{METR Time Horizon.}
\citet{metr_data} released a longitudinal agentic dataset spanning models from GPT-2 (2019) through
February 2026. Two releases are available: TH1.0 covers approximately 33 models through November
2025, and TH1.1 repeats a subset of approximately 20 models with coverage extended to February
2026. We use TH1.0 for the per-band analysis. Task difficulty is labeled by human experts in terms of
expected completion time in minutes, drawing on task sources HCAST, RE-Bench, and SWAA. We treat
completion time buckets as the exogenous difficulty variable.

\paragraph{SWE-bench Verified.}
\citet{swebench} provide 500 software engineering instances, each with a fix time bucket assigned by
humans: under 15 minutes, 15 to 60 minutes, 1 to 4 hours, or more than 4 hours. We collect
per-instance resolved or unresolved outcomes for 134 model submissions from the public experiments
repository~\citep{swebench_experiments}, with model dates recovered from submission folder names.

\paragraph{LiveCodeBench.}
\citet{livecodebench} supply per-problem pass results for $66$ dated models on 1{,}055
problems labeled easy, medium, or hard, with original contest dates that provide contamination
resistance. The outcome for model $m$ on problem $i$ is the number of successes $k_{mi}$ out of
$n_{mi}$ attempts, taken directly from the \texttt{graded\_list} field. We curate these into a
reusable per-item resource, the \emph{LiveCodeBench Difficulty Panel}, recording for every
cell the success count $k_{mi}$, attempt count $n_{mi}$, human difficulty label, and
model release date, which supports analyses stratified by difficulty and by item response beyond the one
in this paper. For most models
$n_{mi} = 10$; for reasoning models $n_{mi} = 1$. Unlike the two agentic benchmarks above,
LiveCodeBench involves direct code generation, a program synthesis task, from a problem statement, with no agentic scaffold.

The $1{,}055$ problems split into $322$ easy, $383$ medium, and $350$ hard items, with pooled base
rates of $0.83$, $0.45$, and $0.18$ respectively, so the hard tail we rely on is well populated
rather than thin. The $66$ models span eleven families (OpenAI $18$, Google $12$, Qwen $10$, Anthropic
$8$, DeepSeek $8$, NVIDIA $3$, Mistral $2$, Moonshot $2$, Meta $1$, xAI $1$, LG $1$), of which $52$ are post-era
and $32$ are trained for reasoning (full roster in \cref{app:roster}). The family sizes are uneven, and OpenAI is the
largest, which is the source of the sensitivity to family clustering in \cref{sec:result-shape}.

\begin{table}[t]
\centering
\caption{Composition of the LiveCodeBench Difficulty Panel. Base rate is the pooled
$\sum k / \sum n$ within each difficulty level.}
\label{tab:panel}
\begin{tabular}{lcc}
\toprule
Difficulty & Items & Base rate \\
\midrule
Easy   & $322$ & $0.83$ \\
Medium & $383$ & $0.45$ \\
Hard   & $350$ & $0.18$ \\
\bottomrule
\end{tabular}
\end{table}

One feature of the panel is collinear with era and deserves comment. The attempt count $n_{mi}$ is
$10$ for almost all pre-era models but $1$ for most post-era models, because submitters sample
reasoning models once. Under the continuous binomial likelihood the attempt count enters as known sampling precision, so
post-era hard-item cells carry more noise.

 Because the count is correlated with era, we test its
role directly rather than assume it is harmless. A simulation that plants a known effect on the real attempt counts shows
the binomial estimator recovers it, while the unweighted estimator (equal weight or a single draw)
is itself biased downward, past zero, and so is not a valid check (\cref{app:nweight}). The small
generated-regressor bias of \cref{app:poscontrol} is present regardless of the weighting and is
calibrated separately. Splitting the post-era models by how often they
were run, the models run ten times and the frontier models run once both show a positive era effect
against the pre-era baseline, which rules out the reading that the effect is confined to the densely
sampled models (\cref{app:nweight}).

\paragraph{Reasoning era breakpoint.}
We define the reasoning era as beginning with o1-preview~\citep{openai_o1}, released in September 2024 (decimal year
2024.70). The binary indicator $\mathit{post}_m$ equals one for all models with a release date at or
after this breakpoint. This cutoff is applied uniformly across all three datasets.

Our estimand is retrospective and structural. We ask whether the transition at this fixed historical
breakpoint changed the shape of the difficulty-response curve, not whether the current frontier
differs from the past. The claim is therefore anchored to the $2024.70$ transition rather than to the
date of analysis, and it does not depend on including the most recent models. A data window that
straddles the breakpoint is what the question requires. The variant in continuous time in
\cref{sec:result-shape} shows the hard-item effect is an ongoing trend within the window, not a
single step.

\subsection{Item response models}
\label{subsec:irt}

Let $y_{mi} \in \{0,1\}$ denote the binary outcome of model $m$ on item $i$, or let $k_{mi}$ of
$n_{mi}$ denote the corresponding binomial counts when multiple attempts are available. We define
model ability $\theta_m \in \mathbb{R}$ and item difficulty $b_i \in \mathbb{R}$, and write
$\mathit{post}_m \in \{0,1\}$ for the post-era indicator and $\mathit{hard}_i \in \{0,1\}$ for a
binary indicator of hard items.

\paragraph{Rasch model (null).}
The one-parameter logistic model serves as the null:
\begin{equation}
  \operatorname{logit} p_{mi} = \theta_m - b_i.
  \label{eq:rasch}
\end{equation}
A model's probability of success on any item is fully determined by the scalar gap $\theta_m - b_i$,
with no structure that depends on era.

\paragraph{Era-DIF model.}
We augment the Rasch model with a single interaction term between era and difficulty:
\begin{equation}
  \operatorname{logit} p_{mi} = \theta_m - b_i + \delta \cdot \mathit{post}_m \cdot \mathit{hard}_i.
  \label{eq:era-dif}
\end{equation}
The parameter $\delta$ is the differential item functioning on hard items that the era introduces: the
additional log odds of success that post-era models achieve on hard items, beyond what their
overall ability $\theta_m$ would predict under the Rasch null. A positive $\delta$ indicates that
the reasoning era has disproportionately improved performance on hard items, relative to easy and
medium items at equal ability.

Throughout, we refer to this separation of a level shift from a hard-item shape change as
CurveShift.

\paragraph{Why not a free 2PL?}
\label{para:no-2pl}
The natural alternative is that hard items simply discriminate differently from easy items, captured
by a two-parameter logistic model, $\operatorname{logit} p_{mi} = \alpha_i \theta_m - \beta_i$, with
per-item discrimination $\alpha_i$. If hard items have higher discrimination, a 2PL with no era term
could in principle absorb the pattern that \cref{eq:era-dif} attributes to $\delta$. However,
jointly estimating per-difficulty discrimination alongside $\delta$ is numerically degenerate on
these datasets: discrimination and DIF are nearly aliased and trade off in the likelihood. The joint
maximum likelihood estimator is unstable across initializations and exhibits sign reversals in
$\delta$ as a function of the starting point. We document this degeneracy in
\cref{app:identifiability}. We therefore do not report a free 2PL control. Instead we adopt an
anchored 2PL sensitivity design, which removes the three channels through which a hard-item era
effect could otherwise be absorbed.

\begin{enumerate}[label=\textbf{Step~\arabic*.}, leftmargin=*, itemsep=4pt]
\item \textbf{Estimate ability from easy and medium items only, then freeze it.}
We fit a binomial likelihood using only easy and medium items to obtain $\hat{\theta}_m$. Hard items
play no role in this estimation. Freezing $\hat{\theta}_m$ closes the primary channel by which a
hard-item DIF signal could be absorbed into the ability dimension: because hard-item outcomes never
enter the ability estimation, any pattern in hard-item residuals cannot be explained away by
adjusting $\theta_m$.
\item \textbf{Fit a model with adjusted discrimination on hard items only.}
With $\hat{\theta}_m$ treated as a fixed offset, we fit
\begin{equation}
  \operatorname{logit} p_{mi} = \alpha_{\mathrm{hard}} \cdot \hat{\theta}_m + c_i + \delta \cdot \mathit{post}_m,
  \label{eq:anchored-2pl}
\end{equation}
where $c_i$ is a per-item fixed effect that absorbs difficulty at the item level, $\alpha_{\mathrm{hard}}$
is a discrimination scaling factor that is pinned rather than estimated, and $\delta$ is the sole
free era parameter. Because $\hat{\theta}_m$ is a frozen offset and $\alpha_{\mathrm{hard}}$ is not
estimated from the data, neither can trade against $\delta$.

\item \textbf{Report $\delta$ across a grid of $\alpha_{\mathrm{hard}}$ values.}
We evaluate \cref{eq:anchored-2pl} at $\alpha_{\mathrm{hard}} \in \{0.55,\, 0.83,\, 1.00\}$. The
value $\alpha_{\mathrm{hard}} = 1.00$ imposes equal discrimination between hard items and the rest
and serves as the conservative headline. The value $\alpha_{\mathrm{hard}} = 0.83$ is a debiased
estimate from the continuous likelihood, and $\alpha_{\mathrm{hard}} = 0.55$ is the deflated
discrimination artifact that arises from binarizing hard items, which have low base rates.
\end{enumerate}

Together, freezing $\hat{\theta}_m$, pinning $\alpha_{\mathrm{hard}}$, and including per-item fixed
effects $c_i$ remove the three channels through which a hard-item era effect could be absorbed:
the ability estimate, item difficulty conflation, and discrimination aliasing. The residual
$\delta$ is therefore identified as a shift across eras on hard items not attributable to any of
these confounds.

\paragraph{Uncertainty quantification.}
We use the continuous binomial likelihood with weights equal to attempt counts instead of
binarizing to pass@1, because binarization deflates discrimination estimates on hard items with low
base rates (\cref{app:identifiability}). Uncertainty is quantified via cluster bootstrap over items,
contests, and model families. Within each bootstrap replicate, model ability $\hat{\theta}_m$ is
re-estimated from easy and medium items before $\delta$ is fit on hard items, so that uncertainty
from treating the frozen ability as a generated regressor propagates into the bootstrap distribution of
$\delta$~\citep{pagan1984generated}. Likelihood ratio $p$-values are reported in \cref{app:lr}
rather than in the main text, because the structure of repeated items and repeated models violates the
independence assumptions underlying conditional $p$-values.

\section{Apparent acceleration is largely scalar}
\label{sec:result-scalar}

\begin{figure}[t]
\centering
\includegraphics[width=0.85\linewidth]{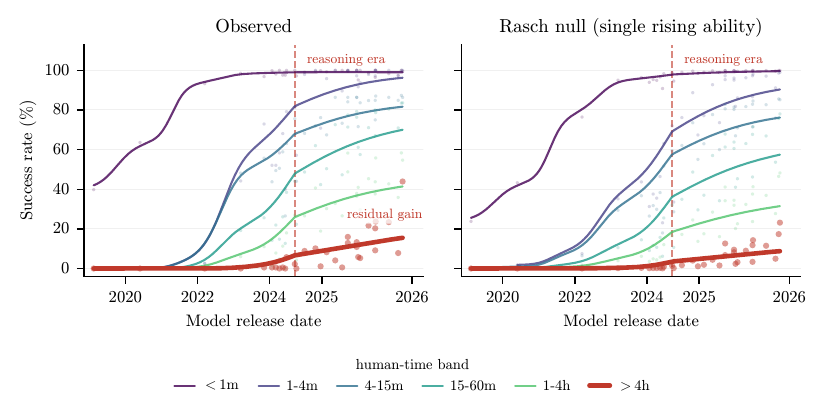}
\vspace{-6pt}
\caption{Success rate versus model release date within each difficulty band on METR Time Horizon, with
difficulty given by exogenous human completion time. Left: observed. Right: predicted by a single
Rasch model with rising ability. The red dashed line marks the reasoning era breakpoint. The time
axis is piecewise linear and is expanded after the breakpoint, so that the reasoning era is legible.
Points are per-model band means and curves are local Gaussian trends through them.}
\label{fig:locus}
\vspace{-8pt}
\end{figure}

A recurring observation in recent agent evaluations is that the fastest improving band of task
difficulty appears to migrate toward harder tasks over time. We refer to this as locus migration.
On the METR Time Horizon data~\citep{metr2025horizon}, with difficulty measured by exogenous human
completion time in minutes, the slope of success against release date within each band shifts markedly between
the eras before and after reasoning (\cref{fig:locus}). \Cref{tab:band-slopes} reports these
slopes. The shortest band stalls near a ceiling, moving from about $10$ to roughly $0$~percentage
points per year (pp/year),
while the longest bands accelerate, with the $15$ to $60$~minute band rising from $6$ to
$37$~pp/year and the over four hour band rising from $0.3$ to $21$~pp/year. From these numbers, this
pattern suggests that the difficulty-response shape itself is changing, with harder tasks
beginning to catch up.

\begin{table}[t]
\centering
\caption{Slopes of success against release date, within each difficulty band (pp/year), on METR Time Horizon, with
difficulty given by exogenous human completion time, for the eras before and after reasoning. The final column is the
after-reasoning slope predicted by the single rising-ability Rasch null; it places the fastest-improving band at $15$ to
$60$ minutes as observed and leaves a positive residual only on the hardest band (\cref{fig:locus}, right).}
\label{tab:band-slopes}
\begin{tabular}{lccc}
\toprule
Difficulty band & Before reasoning & After reasoning & After, Rasch null \\
\midrule
under 1 min   & $10$  & $\sim 0$ & $2.0$  \\
1 to 4 min    & $18$  & $17$     & $31.3$ \\
4 to 15 min   & $12$  & $22$     & $28.7$ \\
15 to 60 min  & $6$   & $37$     & $38.8$ \\
1 to 4 h      & $3.5$ & $27$     & $25.6$ \\
over 4 h      & $0.3$ & $21$     & $12.9$ \\
\bottomrule
\end{tabular}
\end{table}

This reading does not survive a scalar control. We fit a single Rasch model with rising
ability, in which one dimension of ability grows over time and item difficulties are fixed, and we
ask what per-band slopes this model predicts under uniform ability growth. The Rasch null reproduces
the qualitative migration. In particular it places the fastest improving post-era band at the same
difficulty as the observed data, the $15$ to $60$~minute band (\cref{fig:locus},
\cref{tab:band-slopes}). The mechanism is mechanical. Bands near the floor
or ceiling compress their observable slope, and bands of intermediate difficulty sit on the steep
part of the logistic, so uniform growth in a scalar ability is enough to move the apparent locus
toward harder tasks as ability rises. The visual impression that harder tasks are catching up is consistent with a ceiling and discrimination artifact, a form of benchmark saturation, and it is not by itself evidence
that the difficulty-response shape has changed. We read this as a finding in its own right, not only
as a control for what follows: the reported migration of gains is, to first order, the same kind of
measurement artifact that a scalar null produces for emergent abilities~\citep{schaeffer2023mirage},
now on the axis of task difficulty rather than model scale.

The hardest band retains a positive residual relative to the Rasch prediction, the one feature the
scalar null does not fully explain. The residual is small relative to the overall
migration, but it is directionally consistent across the era after reasoning and motivates a targeted
test of whether the difficulty-response shape changes specifically at the hard end. We pursue that
test in \cref{sec:result-shape}.

\section{A hard-item gain survives the scalar control}
\label{sec:result-shape}

To isolate a shape change from a level change, we turn to the no-scaffold
LiveCodeBench~\citep{livecodebench}, where model capability is measured without an agentic harness.
We estimate a hard-item era effect under the anchored 2PL sensitivity design of
\cref{subsec:irt}. Ability $\theta$ is frozen from performance on easy and medium items, items enter
through fixed effects, and the hard-item discrimination $\alpha_{\mathrm{hard}}$ is pinned rather
than freely estimated. A free 2PL is not stably identified on these data because the hard-item
discrimination and the era effect trade off against each other (\cref{app:identifiability}). We
therefore report the era effect across a grid of pinned discrimination values and treat the
equal-discrimination case as the headline. The headline is a hard-item gain of about $+0.40$ logits.
It stays positive under the most conservative discrimination on the grid and through the robustness
checks that follow, its magnitude depends on the model population, and it is carried mainly by the
strongest reasoning series.

\Cref{tab:hard-item-delta} reports the hard-item era effect $\delta$ as a function of
$\alpha_{\mathrm{hard}}$. We report values calibrated for a small positive bias from estimating
ability as a generated regressor; the raw values are larger by at
most $0.05$ logits and appear in the table for reference (\cref{app:poscontrol}). The effect is
positive across the entire grid. At $\alpha_{\mathrm{hard}}=0.55$, the value implied by deflation
under naive binarization, it is large, about $+1.35$ logits (raw $+1.36$). At the debiased value
$\alpha_{\mathrm{hard}}=0.83$ it is about $+0.74$ (raw $+0.78$). At the conservative
equal-discrimination value $\alpha_{\mathrm{hard}}=1.00$, which we adopt as the headline, it is about
$+0.40$ (raw $+0.45$).

We assess its uncertainty with a ladder of cluster bootstraps that correspond to different
generalization targets~\citep{cameron2015cluster}, each shifted by the same calibration. Using $1{,}500$ bootstrap replicates,
clustering by item ($1{,}055$ clusters) gives $[+0.16, +0.67]$, by contest ($\sim 100$ clusters) gives
$[+0.11, +0.70]$, and by model ($66$ clusters) gives $[+0.09, +0.60]$; all three exclude zero, with
$100\%$, $100\%$, and $99\%$ of bootstrap draws positive respectively. The calibration offset is
itself estimated to within about $0.01$, which leaves the lower bound at the model level above zero.
Clustering by model family ($11$ clusters) gives a raw interval of $[-0.25, +0.73]$ that includes
zero; we leave that level uncalibrated, since we draw no inference from it.

Some models appear at more than one reasoning-effort setting, with o1, o3-mini, and o4-mini each run
at low, medium, and high, and Claude Opus 4, Claude Sonnet 4, and Kimi k1.6 each appearing in both a
base and a thinking mode, so their rows are not independent. Pooling every base model into one unit
($66$ to $57$) moves the point estimate only from $+0.45$ to $+0.44$ raw, and clustering by base
model gives a raw interval of $[+0.13, +0.64]$ that still excludes zero, in line with the model level.

We read this ladder as follows. The effect is significant for generalization to new items, new
contests, and new model
checkpoints, the three levels with enough clusters for the cluster bootstrap to be reliable. It is
not individually significant only for generalization to an unseen model family, where the bootstrap
rests on eleven clusters and is underpowered by construction, and where a single influential family,
OpenAI, carries much of the effect (see the leave-one-family-out check below). For the family
level specifically we also run the wild cluster bootstrap of \citet{cameron2008wild}, the standard
remedy when clusters are few; it is marginal and in line with the ordinary family bootstrap
(two-sided $p \approx 0.085$), which indicates the family level is
underpowered, not that the effect is absent there. Our claim is conditional on the existing
population of models.

\begin{figure}[t]
\centering
\includegraphics[width=0.62\linewidth]{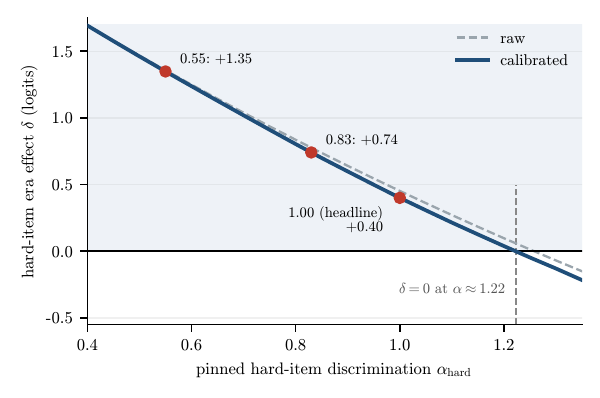}
\vspace{-6pt}
\caption{Hard-item era effect $\delta$ as a function of the pinned discrimination
$\alpha_{\mathrm{hard}}$. The solid curve is calibrated for the generated-regressor bias; the light
dashed curve is raw. The shaded region is $\delta > 0$. The three grid points used in
\cref{tab:hard-item-delta} are marked; the calibrated $\delta$ stays positive until
$\alpha_{\mathrm{hard}} \approx 1.22$, measured on a grid through $\alpha_{\mathrm{hard}}=1.3$ rather
than extrapolated. Values above $1.0$ enter the sweep as adversarial sensitivity points; the
calibrated estimate of hard-item discrimination on these data is $0.83$, and we find no evidence
above one.}
\label{fig:alpha}
\vspace{-8pt}
\end{figure}

Because larger pinned discrimination shrinks the estimated effect, the equal-discrimination headline
is the most conservative point on the grid we consider, and it remains positive.
Sweeping $\alpha_{\mathrm{hard}}$ continuously (\cref{fig:alpha}), the calibrated $\delta$ stays
positive for every value up to $\alpha_{\mathrm{hard}} \approx 1.22$, which we measure rather than
extrapolate, and only then crosses zero. The effect would disappear only if hard items were
about a fifth more discriminating than easy and medium items, whereas the calibrated estimate on these
data is $0.83$, below one, and binarization pushes measured discrimination further down, not up
(\cref{app:binarization}).

\begin{table}[t]
\centering
\caption{Hard-item era effect $\delta$ (logits) on no-scaffold LiveCodeBench under the anchored 2PL
sensitivity design, calibrated for the generated-regressor bias (\cref{app:poscontrol}) with the raw
value alongside. The equal-discrimination row ($\alpha_{\mathrm{hard}}=1.00$) is the conservative
headline. Intervals are item-cluster bootstraps over $1{,}500$ replicates; clustering by base model,
which pools a model's reasoning-effort settings, gives a similar headline interval ($[+0.13, +0.64]$).
This is a sensitivity result, not a free 2PL estimate.}
\label{tab:hard-item-delta}
\begin{tabular}{lccl}
\toprule
$\alpha_{\mathrm{hard}}$ & $\delta$ calibrated & raw & Item-cluster $95\%$ CI \\
\midrule
$0.55$ (deflated by binarization)  & $+1.35$ & $+1.36$ & $[+1.15,+1.58]$ \\
$0.83$ (debiased)              & $+0.74$ & $+0.78$ & $[+0.48,+0.99]$ \\
$1.00$ (equal disc., headline)  & $+0.40$ & $+0.45$ & $[+0.16,+0.67]$ \\
\bottomrule
\end{tabular}
\end{table}

The headline survives a sequence of robustness checks. The values in these checks are raw
$\hat{\delta}$; the calibration of \cref{app:poscontrol} applies to each and shifts it down by less than
$0.08$ at $\alpha_{\mathrm{hard}}=1.0$ (about $0.05$ at the headline), which changes no conclusion.

\paragraph{Anchor sensitivity.}
The choice of which items anchor $\theta$ does not flip the sign. At $\alpha_{\mathrm{hard}}=1.00$,
anchoring $\theta$ on easy items alone gives $\delta \approx +0.72$ logits, anchoring on medium items
alone gives $\delta \approx +0.32$ logits, and anchoring on easy and medium items jointly gives the
headline value (raw $+0.45$). The effect stays positive across all three anchors,
although its magnitude varies, which is expected given that different anchors absorb different
amounts of the signal from easy items.

\paragraph{Separation within the cohort.}
A natural question is whether the effect is specific to reasoning models or merely reflects a 2024
cohort. To test this, we restrict to post-2024 models (32 trained for reasoning and the contemporaneous
non-reasoning remainder of the post-era cohort) and add a term that separates the two groups. The
post-by-hard cohort baseline is $\approx +0.33$ logits, and the excess term specific to reasoning is
$\approx +0.29$ logits: models trained for reasoning carry an additional hard-item gain beyond their
contemporaneous non-reasoning peers. This split within the cohort is exploratory and suggestive, not
confirmatory, and our headline claim keeps the broader post-era cohort framing: on an earlier subset
that undersampled one large model family the same term was slightly negative, and it turns positive
only once the cohort is complete.

\paragraph{Leave-one-family-out.}
The effect does not depend on any single model provider. Dropping each of the eleven model families in
turn (OpenAI, Anthropic, Google, DeepSeek, Qwen, NVIDIA, Meta, Mistral, Moonshot, LG, and xAI), the headline $\delta$ at
$\alpha_{\mathrm{hard}}=1.00$ remains positive, ranging from approximately $+0.15$ (dropping OpenAI)
to approximately $+0.55$ (dropping Qwen). The effect is directionally robust to removing
any single provider, but it is not uniform across providers. Dropping OpenAI alone reduces the
headline from $+0.45$ to $+0.15$, whereas dropping any other family leaves it in the range $+0.44$
to $+0.55$. Much of this drop is compositional, not an effect within OpenAI: the family
fixed effects check below, which identifies $\delta$ from within-family pre/post contrasts only,
keeps a positive point estimate even with OpenAI excluded. The pattern is frontier-led, not a uniform cohort
shift: the hard-item gain is carried disproportionately by the strongest reasoning series, and is
present in the same direction but more weakly in other families. This is also the mechanism behind the wide
interval clustered by family in \cref{tab:hard-item-delta}, since a single influential cluster
dominates the between-family variance.

\paragraph{Family fixed effects.}
The leave-one-family check removes a family's models entirely, mixing between-family and within-family
variation. Family fixed effects identify $\delta$ from within-family pre/post contrasts only, in the
four families that span the breakpoint (Anthropic, DeepSeek, OpenAI, Qwen). This estimator carries a
larger generated-regressor bias than the pooled fit, because the family intercepts absorb more of the
noise from estimating ability; we calibrate it against its own null and report it calibrated, unlike the
raw values used elsewhere in this section (\cref{app:poscontrol}). The calibrated within-family effect
is $\delta_{\mathrm{FE}} \approx +0.52$ with item-cluster interval $[+0.21, +0.76]$ (raw $+0.62$),
larger than the calibrated pooled headline. The pooled OpenAI dependence is then mostly a
between-family compositional effect, not the within-family gain. Splitting the cross-breakpoint
families and calibrating each subset against its own null, the within-family effect is a clear
positive with OpenAI and a weak positive without it, the latter with an item-cluster interval that
includes zero (\cref{app:poscontrol}); the gain is largest in OpenAI, consistent with the
frontier-led pattern. It is a within-family post-era property, not one provider's.

\paragraph{Date perturbation.}
The effect is stable under noise in release dates. Perturbing each release date independently by a
uniform $\pm 0.25$~year and refitting, the headline $\delta$ has mean $+0.45$ and ranges from $+0.34$
to $+0.54$ logits across perturbations, and the sign is preserved in every draw.

\paragraph{Breakpoint location and continuous time.}
The result does not hinge on the exact era cutoff. Sweeping the breakpoint from $2024.3$ to $2025.0$,
the headline $\delta$ is positive for every cutoff at or after $2024.4$, rising monotonically from
$+0.20$ at $2024.4$ to $+0.67$ at $2025.0$, with our $2024.70$ choice near the middle at $+0.45$. At
the earliest cutoff, $2024.3$, almost every model is labeled post-era, the contrast vanishes, and
$\delta \approx 0$, as expected. We also drop the binary split entirely and use an interaction in
continuous time, $\delta \cdot (\mathrm{year}_m - 2024.7) \cdot \mathit{hard}_i$. The recovered hard-item
slope is $+1.03$ logits per year. The hard-item gain accelerates; it does not switch
on at a single date.

\paragraph{Replication across contest sources.}
LiveCodeBench pools problems from two separate contest ecosystems, Codeforces and AtCoder, judged by
the same $66$ models over the same dates. The effect holds on each. Codeforces gives
$\delta \approx +0.79$ with an item-cluster interval of $[+0.40, +1.14]$ on its $453$ problems, and
AtCoder gives $\delta \approx +0.30$ with $[+0.01, +0.59]$ on its $602$. The point estimate is larger
on Codeforces, but the two intervals overlap, so we treat the gap as source heterogeneity rather
than a clean contrast, and the pooled estimate sits between them. The effect is thus not the product
of any single contest source. Both sources are competitive programming, so this is replication
across sources and not across domains.

\paragraph{Contamination filtering.}
Filtering to contamination-safe cells leaves the effect intact and slightly larger. LiveCodeBench
dates its problems by contest, so we keep only cells where the problem postdates the model's
release~\citep{roberts2023contamination}; on this subset (about $15\%$ of cells across $39$ models)
the raw effect is $\delta \approx +0.67$ (the full-sample calibration does not transfer here, the
$n$ geometry differs). Each model is then restricted to problems from its own post-release window, so
the filtered estimate also reflects drift in problem composition over time; we report it as a
robustness check, not a contamination-corrected estimate. The direction is consistent with
contamination inflating pre-era models and shrinking the era contrast, under which reading the
headline is conservative.

Across the main robustness checks the sign is preserved; only the magnitude moves. A nonparametric Mantel-Haenszel check,
which stratifies on ability instead of fitting any logistic form, agrees in direction and magnitude
(\cref{app:mh}). The estimate is a sensitivity result, not
a free 2PL fit. The free model is not stably identified on these data, so we pin discrimination and report
the most conservative point on the grid that we can defend. A positive control simulation confirms
that the anchored design recovers a planted effect with slope near one, and that its null bias is a
small positive value, at most $+0.08$, far below the effect and corrected for by the calibration, so the
positive $\delta$ is not manufactured by the procedure (\cref{app:poscontrol}).

\section{Only a no-scaffold benchmark separates capability from scaffold}
\label{sec:result-identification}

The hard-item residual is not unique to LiveCodeBench. The same directional effect appears in
agentic benchmarks, but there it cannot be cleanly attributed to model capability, because model era
and scaffold era are collinear. This is why a no-scaffold benchmark is
needed (\cref{tab:triangulation}).

The collinearity is concrete in the METR data. In the hardest band, if we restrict to a single
scaffold that spans both eras, only one post-era model remains, and that model is not itself
trained for reasoning. Post-era models are essentially never run under harnesses from before the
reasoning era, so era and harness move together and cannot be told apart.
SWE-bench~\citep{swebench} shows the same structure. Its
directional hard-item effect is positive but not statistically significant, and it suffers from both
a thin hard tail and its own scaffold era collinearity, especially because agentic and agentless
solution procedures can themselves change SWE-bench outcomes~\citep{yang2024sweagent,xia2024agentless}.

LiveCodeBench breaks the collinearity by construction. It has no agentic scaffold, so there is no
harness era to confound with the model era. The hard-item era effect on LiveCodeBench isolates deployed model capability, including native inference-time reasoning, from the contribution
of any harness. The result of \cref{sec:result-shape} can therefore be read as a statement
about models, not scaffolds.

The properties that make LiveCodeBench usable here are rare in combination. A benchmark must pair
dated models that straddle the reasoning era breakpoint with per-item outcomes, an exogenous
difficulty ordering, and the absence of an agentic scaffold. Few public benchmarks meet all four requirements
at once. The nearest alternatives each fail one condition: CodeElo~\citep{quan2025codeelo} 
shares the Codeforces source rather than adding an independent one, LiveBench~\citep{white2024livebench} lacks a clean exogenous
per-item difficulty, practical code-generation suites such as
BigCodeBench~\citep{zhuo2024bigcodebench} do not provide the same dated no-scaffold panel with contest
difficulty, and dated math competition sets such as Omni-MATH~\citep{gao2024omnimath} do not publish a per-item
matrix across dated models.

We tried one such math panel directly, and the way it fails is itself informative. On the MathArena
2025 contests~\citep{balunovic2025matharena}, models answer without a scaffold and the contest origin gives a difficulty ordering,
so two of the four conditions are met. The other two are not. The only per-item difficulty is the
problem index, an endogenous proxy rather than an exogenous ordering, and the era coverage is nearly
absent: these panels evaluate the models available when each contest is released, so the pre-era
side holds zero to two models per contest. With the pre-era hard-item cells almost empty, the
post-by-hard contrast has no pre-era support, and the era coefficient is not defined on these data
before any model is fit. A fit run anyway returns a degenerate estimate near zero, which reflects
this missing support rather than an absence of the effect (\cref{app:matharena}). The failure is not
a benchmark that disagrees; it is a benchmark that cannot pose the question.

The reliance on a single benchmark reflects the current data landscape, and the
window is closed in retrospect: the pre-era frontier endpoints are no longer served, so the era contrast
cannot be collected on a new benchmark after the fact.

\Cref{tab:triangulation} summarizes the three sources. METR, an agentic benchmark, shows a hard-item
effect of $+9.0$~pp/year that is significant but confounded by scaffold era. SWE-bench, also
agentic, shows $+2.3$~pp/year that is not significant and is likewise confounded. LiveCodeBench, with
no scaffold, shows a positive and significant effect. The agentic sources establish that the
directional pattern is present where it matters most for deployment, and the no-scaffold source
establishes that at least part of the pattern reflects model capability rather than harness.

\section{Discussion}
\label{sec:discussion}

\paragraph{Aggregate progress curves conflate a level shift with a shape change.}
Popular summaries of agent progress compress capability onto one axis. METR time horizons map
ability to the task duration a model can complete~\citep{metr2025horizon}, and Epoch's latent ability
stitching aligns heterogeneous benchmarks onto a shared scale~\citep{epoch_rosetta}. Both are scalar.
The difficulty at which the median model succeeds is rising, and this is real, but
the same average trend can arise from very different redistributions of difficulty. A uniform upward
shift of an ability parameter, a steepening of item discrimination near a ceiling, and a genuine
change in how models respond to the hardest items are not distinguishable at the level of a single
aggregate curve. Our results show that the reasoning era did more than move the level. After the
scalar component is removed, the hard-item difficulty-response shifts upward beyond what rising
ability predicts. The story remains scalar to first order, which is why the Rasch
model~\citep{rasch1960} reproduces the widely reported migration, but a shape change at second order is present
and survives a conservative test.

\paragraph{Naive psychometric tests mislead in opposite directions.}
A Rasch null leaves no room for a hard-item residual, so it pushes the visible shape change onto a
ceiling artifact: its single ability parameter forces every era effect through the level. A free
2PL makes the opposite mistake. Because discrimination and the
interaction of era with difficulty are not stably identified jointly on this data, fitting both lets estimated
discrimination absorb the very signal the test is meant to detect, biasing the residual toward
zero~\citep{holland1993dif, birnbaum1968}. Two further checks mislead in the same direction: dropping
the weighting by attempt count biases the estimate downward, past zero, and even the correct estimator
carries a small, measurable generated-regressor bias
(\cref{app:nweight}, \cref{app:poscontrol}).

A design that avoids these failure modes anchors ability
on the easy and medium items, adds item fixed effects, reports a sensitivity grid over pinned discrimination
instead of a single free value, and calibrates for the residual bias instead of treating it as a
patch. The hard-item era effect we report is the value that survives the least favorable point on
that grid, the equal-discrimination assumption, so it should be read as the most conservative value
on the defensible grid, not a point estimate.

\paragraph{A possible mechanism.}
The pattern is consistent with reasoning training~\citep{deepseek_r1}, chain-of-thought-style reasoning~\citep{wei2022cot}, and native test-time compute~\citep{snell2024testtime} helping most where
verification is available, since contest items are hard but checkable and reward additional
inference-time search~\citep{yao2023treeofthoughts} and filtering~\citep{li2022alphacode,cobbe2021verifiers,lightman2023verify}. We did not test this mechanism, and our design cannot isolate it. The era
label bundles reasoning training with concurrent scaling and test-time compute, so we do not
attribute the effect to reasoning training alone. The cheapest compute account fails a direct
check: pre-era models still trail on hard items when granted their full ten-attempt sampling
budget (\cref{app:budget}). The pattern is also frontier-led rather than
uniform: it is carried disproportionately by the strongest reasoning series. Removing that one
provider shrinks the pooled headline, but a within-family analysis with fixed effects shows most of that
shrinkage is compositional, and the within-family point estimate remains positive, though imprecise,
when that provider is excluded (\cref{sec:result-shape}). Our comparison within the cohort offers suggestive evidence, dependent on the sample,
that models trained for reasoning carry a hard-item excess beyond their contemporaries; because
this split flips sign with cohort composition, we treat it as exploratory and do not rest the
headline claim on it.

\paragraph{Implications for measurement.}
This suggests two changes to reporting practice. Progress stratified by difficulty should be reported
alongside any scalar summary, because a scalar cannot separate a level shift from a shape change and
obscures redistributions of difficulty that matter for capability forecasting. And on agentic benchmarks, where
model era and scaffold era are collinear, scaffold era should be treated as a potential confound and
no-scaffold measurements preferred when the claim concerns the model. We do not claim this resolves
autonomy over long horizons, where scaffolds are integral to the task and the no-scaffold counterfactual is
not well defined. A time-horizon doubling rate~\citep{metr2025horizon} summarizes only the level; the
same doubling can come from uniform progress or from a shifting hard tail, and reporting it by
difficulty band would tell the two apart.
It helps to read observed performance as a model term, a scaffold term, their interaction, and a task term. Our design identifies the model term by removing the scaffold, while the interaction between model and scaffold stays unidentified wherever model era and scaffold era are collinear. We use this decomposition to mark that boundary, not as a general estimator we can fill in here.

\section{Limitations}
\label{sec:limitations}

Our identification is established on LiveCodeBench contest programming: short horizon tasks with
deterministic verifiers, no agentic scaffold, and difficulty that comes from reasoning depth on
fully specified problems, not from gathering information. The effect replicates across its two contest sources, but both are competitive programming, so
this is replication across sources, not domains. Few public benchmarks combine dated models, exogenous difficulty labels,
per-item outcomes, and no scaffold; a no-scaffold mathematics benchmark with a comparable
difficulty ordering is the natural next target. We tried one already, and it fails on two of these
conditions before a test can be run (\cref{app:matharena}); making it work would require obtaining a
human solve rate per problem as the exogenous anchor, which is a separate data collection effort we
leave to future work. The results do not
extend to agentic benchmarks over long horizons such as METR, SWE-bench, WebArena, or OSWorld~\citep{swebench,zhou2023webarena,xie2024osworld}, where scaffolds and models
coevolve and confound any comparison at the era level; claims about post-era gains there should be
read with caution. In that regime inference-time scaffolding such as agent skills becomes part of the task, not a confound to remove, and measuring capability there is itself an open problem~\citep{skillsrecipe}.

The post-era indicator captures a cohort of releases, not a single intervention: it bundles
training specific to reasoning, continued scaling, native test-time compute, and agent-level
test-time scaling methods~\citep{snell2024testtime,zhu2025agentttc}, and we cannot isolate
any one as the driver. Our comparison of reasoning models against their non-reasoning
contemporaries finds a positive excess specific to reasoning that depends on the sample and flips sign
under incomplete cohorts, so we treat it as exploratory. The estimand throughout is the performance
of a deployed model as evaluated, including any native inference-time reasoning; it is not a pure
effect of model weights, and an intervention on the deployment configuration, such as disabling
chain of thought or matching inference compute across models~\citep{brown2024monkeys}, would
correspond to a different estimand.

On the estimation side, three cautions apply. Joint estimation of $\alpha_{\mathrm{hard}}$ and
$\delta$ is degenerate (\cref{app:identifiability}), so we pin $\alpha_{\mathrm{hard}}$ over a grid
and lead with the conservative equal-discrimination value; a fully Bayesian hierarchical 2PL with a
shrinkage prior on discrimination would be more efficient; we leave it to future work. The
post-era indicator depends on release dates coded by hand, on which we report a date perturbation
check (\cref{sec:result-shape}); the analysis inherits any systematic error in them. And many
specifications were explored before the design settled, so the conditional likelihood ratio
$p$-values should be read with that search in mind, and the positive result assessed through the
cluster bootstrap intervals and the leave-one-family-out analysis.

\section{Conclusion}
\label{sec:conclusion}

Scalar summaries of agent progress cannot tell a level shift from a change
in the shape of the difficulty-response curve, and a single model with rising ability reproduces the
widely reported migration of gains toward harder tasks. A shape change is still there once the scalar part
is removed, about $+0.40$ logits on the hardest LiveCodeBench items. Post-era models gain on the hardest items beyond their easy and medium ability, and
the gain holds under our most conservative discrimination assumption. It is clearly visible only without an agentic scaffold, where model capability is not
confounded with harness, and we claim it for hard tasks that need only short reasoning, not for autonomy
over long horizons. We offer this as a correction to scalar progress narratives, not as a claim
about what causes the gain.

\section{Reproducibility}
\label{sec:repro}
We release the \emph{LiveCodeBench Difficulty Panel} and all analysis code at
\url{https://github.com/harvenstar/CurveShift}. The panel records, for every cell, the success and attempt
counts $k_{mi}$ and $n_{mi}$, the human difficulty label, and the model release date; the $66$-model roster
is in \cref{app:roster}. All reported estimates come from one pipeline: ability is
frozen from easy and medium items, item fixed effects and the era term are fit on hard items, and
the hard-item discrimination is pinned over the grid $\{0.55, 0.83, 1.0\}$. Cluster bootstrap
intervals use $1{,}500$ replicates with seed $2026$ at the item, contest, model, and family levels. The generated-regressor calibration of \cref{app:poscontrol} is measured by
planting effects on the real attempt counts over $200$ replicates with seed $2026$, on the grid
$\alpha_{\mathrm{hard}} \in \{0.55, 0.83, 1.0, 1.1, 1.2, 1.3\}$ and $\delta_{\mathrm{true}} \in \{0,
0.2, 0.45, 0.7\}$. The family fixed effects estimator of \cref{sec:result-shape} is calibrated
against its own null over $400$ replicates with seed $7$, the cross-breakpoint subsample estimators
the same way over $200$.

\bibliographystyle{tmlr}
\bibliography{references}

\appendix
\section{Identifiability of the free 2PL and the bias of a naive residual test}
\label{app:identifiability}

This appendix documents the methodological pitfalls we encountered and ruled out, so that reviewers
can verify that the obvious alternatives were considered and rejected for principled reasons.

\subsection{The free 2PL is not stably identified jointly with the era DIF coefficient}
\label{app:nonident}

The two-parameter logistic model posits
$\operatorname{logit} p_{mi} = \alpha_i (\theta_m - \beta_i)$, with discrimination $\alpha_i$. The
era DIF hypothesis adds a coefficient $\delta$ for hard items,
\begin{equation}
  \operatorname{logit} p_{mi} = \alpha_{\mathrm{hard}} (\theta_m - \beta_i)
    + \delta \cdot \mathit{post}_m \cdot \mathit{hard}_i.
  \label{eq:dif-app}
\end{equation}
The parameters $\alpha_{\mathrm{hard}}$ and $\delta$ are not stably identified jointly on our data. A higher
$\alpha_{\mathrm{hard}}$ increases the predicted advantage of high ability models on hard items.
Because post-era models tend to have higher fitted ability, raising $\alpha_{\mathrm{hard}}$ already
predicts that post-era models will do better on hard items, and the model can offset this with a
more negative $\delta$. Conversely, a lower $\alpha_{\mathrm{hard}}$ creates room for a larger
positive $\delta$. The two parameters trade off continuously, and the likelihood surface is nearly flat
along a shallow ridge in the $(\alpha_{\mathrm{hard}}, \delta)$ plane: the data carry little
information along it, because post-era ability and the hard-item indicator are strongly but not
perfectly collinear.

This near-degeneracy is not a consequence of optimizer failure. In our joint maximum likelihood
estimation, the sign of $\hat{\delta}$ flips depending on parameterization and initialization, while
$\hat{\alpha}_{\mathrm{hard}}$ remains unstable across random initializations. Such behavior is the
empirical signature of weak identification along a ridge, not of a local optimum. We therefore do not
report a free 2PL fit, and the main text instead pins $\alpha_{\mathrm{hard}}$ over a prespecified
grid and reports how $\hat{\delta}$ varies across it.

\subsection{A residual test from the null 2PL is biased toward zero}
\label{app:residual-bias}

A tempting alternative is to fit a null 2PL with no era term jointly on all data, compute residuals,
and check whether the post-era hard-item cells have systematically positive residuals. We show that
this procedure has near zero power against the true DIF, because fitting the null on data that
contains the true DIF launders the effect into other parameters.

Suppose the data were generated under \cref{eq:dif-app} with a true $\delta = +0.5$ logits. When the
null 2PL is fit to these data, three parameters absorb the signal. Post-era models acquire inflated
fitted ability $\hat{\theta}_m$, because the model attributes the excess performance on hard items
to higher overall ability. Hard items acquire deflated fitted difficulty $\hat{\beta}_i$. Hard items
also acquire inflated fitted discrimination $\hat{\alpha}_i$. In our simulation under true
$\delta = +0.5$, the residual procedure recovered only $+0.016$ logits; the effect was almost
entirely absorbed. A tight bootstrap interval around this near zero estimate provides false
reassurance: the interval is narrow precisely because the bias is systematic, not because the DIF is
absent. A residual estimate near zero, or even slightly negative, is therefore fully consistent with
a genuine positive DIF of substantial magnitude. This is why our main analysis freezes ability
estimation on easy and medium items and pins discrimination, closing both absorption channels.

\subsection{Binarization deflates discrimination on hard items}
\label{app:binarization}

One might binarize each cell at $\mathrm{pass@1} \geq 0.5$ and fit a 2PL to the resulting
binary matrix. We show by simulation that this distorts estimated discrimination, especially on hard
items. Hard items have low base rates, so most binary outcomes are zero regardless of ability.
Because binary success barely covaries with ability under this floor effect, the estimated item
response function has a shallow slope, producing spuriously low discrimination. In our simulation
with true discrimination $\alpha = 1.0$, binarization recovered $\hat{\alpha} = 0.49$, whereas the
continuous binomial likelihood recovered $\hat{\alpha} = 0.83$. The deflation is worst precisely on
hard items, whose discrimination we most need to characterize correctly. We therefore use the
continuous binomial likelihood throughout, and include the discrimination implied by binarization
($\hat{\alpha} \approx 0.55$) only as the lower, conservative end of our sensitivity grid.

\subsection{Calibrating the estimator for a generated-regressor bias}
\label{app:poscontrol}

The negative control above shows the naive residual test is biased toward zero. The symmetric worry
is the opposite failure: that the anchored design manufactures a positive $\delta$ when none is
present. We test this with a positive control and find a small positive bias, which we then measure
and remove.

The pipeline estimates ability from easy and medium items and carries it into the hard-item fit as a
fixed offset, so $\hat{\theta}_m$ is a generated regressor~\citep{pagan1984generated,murphy1985twostep}. To measure its
effect we plant a known effect on the real data: we take item difficulties from a real fit, set
$\operatorname{logit} p_{mi} = \alpha_{\mathrm{hard}} \theta_m + c_i + \delta_{\mathrm{true}}\,
\mathit{post}_m$ on the hard cells and the matching one-parameter form on the easy and medium cells,
draw $k^{*}_{mi} \sim \mathrm{Binomial}(n_{mi}, p_{mi})$, and refit the whole pipeline, including the
re-estimation of $\theta$. \Cref{tab:poscontrol} reports the recovered $\hat{\delta}$ over $200$
replicates.

\begin{table}[t]
\centering
\caption{Positive control under the pipeline we run, with ability re-estimated in each replicate.
Planted $\delta_{\mathrm{true}}$ against the recovered value, mean over $200$ simulations. The
estimator tracks the planted value with slope near one and a small positive offset under the null
that grows with $\alpha_{\mathrm{hard}}$, the signature of a generated-regressor bias.}
\label{tab:poscontrol}
\begin{tabular}{lcccc}
\toprule
$\alpha_{\mathrm{hard}}$ & $\delta_{\mathrm{true}}=0$ & $0.20$ & $0.45$ & $0.70$ \\
\midrule
$0.55$ & $+0.044$ & $+0.235$ & $+0.482$ & $+0.726$ \\
$0.83$ & $+0.076$ & $+0.262$ & $+0.501$ & $+0.739$ \\
$1.00$ & $+0.078$ & $+0.263$ & $+0.498$ & $+0.736$ \\
\bottomrule
\end{tabular}
\end{table}

The estimator tracks $\delta_{\mathrm{true}}$ with slope near one, about $0.94$ at
$\alpha_{\mathrm{hard}}=1.0$ with a standard error of about $0.007$. Under the null it returns a small
positive value, which grows with $\alpha_{\mathrm{hard}}$ from $+0.04$ at $0.55$ to $+0.08$ at $1.0$,
as expected because the generated-regressor noise enters the hard-item fit scaled by
$\alpha_{\mathrm{hard}}$. An earlier version of this control froze $\theta$ at its true value instead
of re-estimating it; that simpler design removes the generated-regressor channel and returns a small
negative offset, but it does not match the pipeline we run, so we calibrate against the faithful
version.

We map each reported estimate by $\hat{\delta}_{\mathrm{cal}} = (\hat{\delta}_{\mathrm{raw}}
- \mathrm{offset}(\alpha_{\mathrm{hard}}))/\mathrm{slope}$ and shift the cluster bootstrap intervals
by the same map; the offset is itself estimated to within about $0.01$. At the headline
$\alpha_{\mathrm{hard}}=1.0$ this takes the raw $+0.45$ to a calibrated $+0.40$. The offset declines
for $\alpha_{\mathrm{hard}}>1$ rather than continuing to grow, so we measure the calibrated zero
crossing on a grid through $\alpha_{\mathrm{hard}}=1.3$ instead of extrapolating, and it sits at
$\alpha_{\mathrm{hard}} \approx 1.22$. The calibrated headline of $+0.40$ is a positive effect that
the procedure does not manufacture: the null bias of the pooled estimator is at most $+0.08$, well
below the effect, and the calibration corrects for it.

The same generated-regressor channel appears in the family fixed effects estimator of
\cref{sec:result-shape}. The added family intercepts enlarge it: planting the null there returns an
offset of about $+0.15$, above the pooled $+0.08$, with the same slope near one ($0.92$). The channel
does not transfer across subsets, and each family split is calibrated against its own null. The four
cross-breakpoint families have offset $+0.24$ and the three without OpenAI $+0.06$; after calibration
the within-family effect is $+0.56$ with OpenAI (raw $+0.77$) and $+0.19$ without it ($26$ models, raw
$+0.25$). The value with OpenAI tracks the full-sample $+0.52$. The value without OpenAI is a weak positive
whose item-cluster interval includes zero ($[-0.10, +0.51]$ calibrated), and we read it as directional.

Like the leave-one-family check, each subset re-estimates ability on its own members; freezing ability
from all $66$ models instead reads the effect without OpenAI as $+0.35$ raw, $+0.29$ calibrated with
interval $[0.00, +0.61]$, still a weak positive touching zero at the lower end, but that mixes a full-sample anchor
with a subset fit and departs from how every other check at the family level in this paper is computed.

\subsection{The weighting by attempt count does not bias the estimate}
\label{app:nweight}

The attempt count $n_{mi}$ is tied to era. Pre-era models are run ten times per problem and most
post-era models once, and under the continuous binomial likelihood the count acts as a weight. A
reader may worry that the era effect is an artifact of this weighting. We check it directly, and we
first explain why one natural check is misleading here.

That check drops the weighting, either by reducing each cell to one draw or by fitting the raw
proportions $k_{mi}/n_{mi}$ with equal weight. Both make the headline vanish: the equal weight fit
returns $\delta \approx -0.05$. We initially read this as evidence against the effect; the simulation
below shows that reading was wrong, because the equal weight estimator is itself biased on this
sampling geometry (\cref{tab:nweight}, top). When we plant a known effect on the real
attempt counts, the binomial estimator recovers it with slope near one, while the equal weight
estimator has slope near $0.6$ and an offset near $-0.4$: it returns $-0.09$ when the truth is
$+0.45$ and $-0.38$ when the truth is zero. A fit calibrated this way produces a value near $-0.05$
for a true effect close to $+0.45$, matching the $-0.05$ observed on the real data.

A constant planted effect leaves one question open. The same pair of values would also appear if the
effect were confined to the models run ten times and absent in the frontier models run once. We separate
the two and estimate the effect within each, against the same pre-era baseline (\cref{tab:nweight},
bottom). On the real data the models run ten times give $\delta = +0.43$ with $[+0.18, +0.72]$, and
the frontier models run once give $+0.81$ with $[+0.55, +1.09]$. The second value is inflated. On
the same simulation this slice estimator carries an outward small sample bias of $+0.26$ to $+0.31$
across the calibration range (\cref{tab:nweight}, middle); using the larger offset at the null of
$+0.31$, the corrected value is $+0.50$ with an interval of $[+0.24, +0.78]$. Both groups stay
positive and both intervals exclude zero. The frontier models carry the effect rather than diluting
it, which closes off the alternative that it is concentrated where sampling is dense. The weighting
changes how precisely we estimate the effect, not its sign or rough size, and the binomial estimate,
after the calibration of \cref{app:poscontrol}, is the one to report.

\begin{table}[t]
\centering
\caption{Diagnostics for the attempt count concern. Top: a simulation that plants a known effect on
the real attempt counts; the binomial estimator recovers it while the equal weight estimator is
biased toward negative values. Middle: on the same simulation the binomial estimator is unbiased for
the models run ten times but carries an outward bias of $+0.26$ to $+0.31$ across the calibration
range for the models run once. Bottom: on the real data both groups give a positive era effect
against the pre-era baseline; the row for models run once is shown before and after the bias correction.}
\label{tab:nweight}
\begin{tabular}{lcc}
\toprule
Planted $\delta_{\mathrm{true}}$ & Binomial & Equal weight \\
\midrule
$+0.00$ & $+0.07$ & $-0.38$ \\
$+0.20$ & $+0.26$ & $-0.26$ \\
$+0.45$ & $+0.50$ & $-0.09$ \\
$+0.70$ & $+0.73$ & $+0.07$ \\
\midrule
Planted $\delta_{\mathrm{true}}$ & $n{=}1$ slice & $n{=}10$ slice \\
\midrule
$+0.00$ & $+0.31$ & $+0.05$ \\
$+0.45$ & $+0.71$ & $+0.48$ \\
\midrule
Real data subgroup & $\delta$ & $95\%$ CI \\
\midrule
post, $n{=}10$                  & $+0.43$ & $[+0.18, +0.72]$ \\
post, $n{=}1$ (raw)             & $+0.81$ & $[+0.55, +1.09]$ \\
post, $n{=}1$ (corrected)       & $+0.50$ & $[+0.24, +0.78]$ \\
\bottomrule
\end{tabular}
\end{table}

\section{Likelihood ratio statistics}
\label{app:lr}

For completeness we report the conditional likelihood ratio statistics for the anchored era effect.
Under the anchored design with $\alpha_{\mathrm{hard}}$ pinned and ability frozen, the era term is
highly significant at every grid point. The nested comparison adds the single era parameter $\delta$
to the null, so the statistic has one degree of freedom; the fit with a warm start yields $\mathrm{LR}
\approx 85$ at $\alpha_{\mathrm{hard}}=1.0$, far beyond the $\chi^2_1$ critical value of $3.84$. We
do not treat these $p$-values as the
primary evidence, because repeated items and repeated models violate the independence assumptions
behind the conditional likelihood, which makes the resulting $p$-values anticonservative. The
cluster bootstrap intervals in \cref{sec:result-shape} are the inferential basis for our claims.

\section{A nonparametric check}
\label{app:mh}

Our inference rests on a logistic family, so we add one check that does not. A Mantel-Haenszel~\citep{mantel1959mh}
analysis matches models on their easy and medium ability, then compares pre-era and post-era
performance on hard items within ability strata, with no functional form. The common odds ratio is
$1.84$ (log odds $+0.61$), favoring post-era models, in the direction and rough magnitude of the
parametric estimate. The matching variable overlaps only in the lower ability strata, where pre-era
and post-era models coexist; the comparison is identified there. As with the likelihood ratio
statistics, we read the odds ratio, not its $p$-value, which is anticonservative under repeated
items and models.

\section{A budget handicap check}
\label{app:budget}

Post-era models spend more inference compute per attempt, and repeated sampling alone is known to
buy coverage on hard problems~\citep{chen2021codex,wang2022selfconsistency,brown2024monkeys}. A natural question is how much of the
hard-item era effect is within reach of simply giving pre-era models more attempts. The panel
cannot price tokens (\cref{sec:limitations}), but it does contain a budget axis: pre-era models
were run ten times per problem. We rerun the headline contrast with the pre-era hard-item outcome
replaced by success within the available budget, $1\{k_{mi} \geq 1\}$, against the unchanged
post-era outcomes. The ability anchor stays on the per-attempt scale, because coverage at ten
attempts saturates the easy items.

The empirical facts come first. On hard items, pre-era models solve $3.8\%$ of cells per attempt
and reach only $7.0\%$ within ten attempts, while post-era models solve $27.6\%$ in a single
attempt. Ten draws buy less than a doubling because the attempts are strongly correlated: the
intra-cell correlation is about $0.51$ on pre-era hard cells and $0.76$ on easy and medium cells.
Failures on these items are systematic, not sampling noise.

The model-based version needs its own null, and the first one we built was wrong. Simulating the handicap with independent attempts implies a pre-era coverage of $19.6\%$ where
the data show $7.0\%$; that null over-credited the handicap and inflated the calibrated effect to
$+1.7$, a number we rejected because the simulation could not reproduce the observed statistic. A
beta-binomial null at the correlations estimated per stratum reproduces the observed coverage and
is the calibration we report: planted per-attempt effects on the grid $\{0, 0.45, 0.9, 1.35\}$,
$150$ replicates with seed $7$, inverted within the simulated bracket, never extrapolated. The
same beta-binomial noise leaves the calibration of \cref{app:poscontrol} essentially unchanged
(null offset $+0.077$ against the binomial $+0.078$); the headline does not rest on the
independence assumption either.

Under this null the handicapped era effect is $\delta_{\mathrm{b10}} \approx +0.57$ with mapped
item-cluster interval $[+0.35, +0.78]$; treating every hard cell as coverage at its available
budget instead gives $+0.36$ with $[+0.15, +0.57]$. Both exclude zero. Granting the pre-era cohort
its full tenfold sampling budget on hard items does not absorb the era effect. The check speaks
only to budgets up to ten attempts, one decade of the scaling reported by
\citet{brown2024monkeys}, and to repeated sampling, not to longer reasoning within an attempt.

\section{Cross-source triangulation table}
\label{app:triangulation}

\Cref{tab:triangulation} collects the cross-source comparison discussed in \cref{sec:result-identification}.

\begin{table}[h]
\centering
\caption{Cross-source triangulation of the hard-item era effect. Agentic benchmarks (METR,
SWE-bench) carry model era and scaffold era collinearity, so their effects cannot be attributed to
model capability. The no-scaffold benchmark (LiveCodeBench) has no harness to confound, so its
hard-item effect isolates deployed model-side capability from agentic harness improvements.
Significance for the LiveCodeBench row is at the item, contest, and model cluster levels; clustering
by model family is underpowered (\cref{sec:result-shape}). Units for the agentic rows are pp/year; the
no-scaffold row reports the anchored 2PL headline effect in logits.}
\label{tab:triangulation}
\begin{tabular}{llll}
\toprule
Source & Scaffold & Hard-item effect & Identification \\
\midrule
METR          & Agentic & $+9.0$ pp/yr (significant)        & Confounded by scaffold era \\
SWE-bench     & Agentic & $+2.3$ pp/yr (not significant)    & Confounded, thin hard tail \\
LiveCodeBench & None    & $\approx +0.40$ logits (significant) & Isolates model-side capability \\
\bottomrule
\end{tabular}
\end{table}

\section{A math-domain attempt that fails on the preconditions}
\label{app:matharena}

We tried to move the design to a second domain and report the attempt here, because how it fails is
itself evidence for the identification argument in \cref{sec:result-identification}. The natural
candidate is competition mathematics without an agentic scaffold, where models answer directly and
the contest origin gives a difficulty ordering. We used the MathArena panels for the 2025 contests~\citep{balunovic2025matharena} 
(AIME, HMMT February, BRUMO, SMT, and CMIMC). The attempt does not fail a hypothesis test; it fails
before one can be run, on two of the four conditions that make LiveCodeBench usable.

The first missing condition is a clean exogenous difficulty. The only per-item ordering these
panels carry is the problem index, which is an endogenous and noisy proxy. A genuinely exogenous
difficulty would come from the human solve rate, which the panels do not publish, and using the
model solve rate instead would reintroduce the circularity our design exists to avoid. The second
missing condition is era coverage. These panels evaluate the models available when each contest is
released, so pre-era models are almost absent from the 2025 problems (\cref{tab:matharena}). Because
the pre-era side of the hard-item cells is nearly empty, the post-by-hard contrast has no pre-era
support, and the era coefficient is not defined on these data before any model is fit. A fit run
anyway returns a degenerate estimate, an interaction near $+0.001$, which reflects this rank
deficiency rather than an absence of the effect.

\begin{table}[h]
\centering
\caption{Era coverage of the MathArena 2025 panels, by contest. Pre-era counts models released
before the $2024.70$ breakpoint. The pre-era column is nearly empty, so the post-by-hard era
contrast is not identified on these panels.}
\label{tab:matharena}
\begin{tabular}{lccc}
\toprule
Contest & Models & Pre-era ($<2024.70$) & Post-era \\
\midrule
HMMT February 2025 & $64$ & $2$ & $62$ \\
AIME 2025          & $67$ & $2$ & $65$ \\
BRUMO 2025         & $44$ & $0$ & $44$ \\
SMT 2025           & $43$ & $0$ & $43$ \\
CMIMC 2025         & $35$ & $0$ & $35$ \\
\bottomrule
\end{tabular}
\end{table}

The window that would let a math panel work is closed in the same way as elsewhere. The pre-era
contrast would need frontier models from before the breakpoint, and those closed endpoints are
retired. Open-weight pre-era models remain and could in principle seed such a panel, but they are
underpowered for this contrast and do not include the providers that carry the effect. We therefore
treat a no-scaffold math benchmark with an exogenous per-item difficulty as future work, not as a
comparison available now.

\section{Model roster}
\label{app:roster}
The $66$ LiveCodeBench models, with hand-coded release dates (decimal year), family, era, and
reasoning label. Models at or after $2024.70$ are post-era.\footnote{The (Low), (Med), and (High)
entries are reasoning-effort settings of one underlying model, and a few systems appear in both a
base and a thinking mode, so these rows are not independent. We report a robustness check that pools
them to one unit per base model in \cref{sec:result-shape}.}
{\small
\begin{longtable}{llllc}
\toprule
Model & Date & Family & Era & Reasoning \\
\midrule\endhead
GPT-4-0613 & 2023.45 & OpenAI & pre & no \\
GPT-4-Turbo-1106 & 2023.85 & OpenAI & pre & no \\
DSCoder-6.7b-Ins & 2023.86 & DeepSeek & pre & no \\
DSCoder-33b-Ins & 2023.86 & DeepSeek & pre & no \\
DSCoder-1.3b-Ins & 2023.86 & DeepSeek & pre & no \\
Claude-3-Haiku & 2024.20 & Anthropic & pre & no \\
GPT-4-Turbo-2024-04-09 & 2024.27 & OpenAI & pre & no \\
GPT-4O-2024-05-13 & 2024.37 & OpenAI & pre & no \\
Codestral-Latest & 2024.42 & Mistral & pre & no \\
Qwen2-Ins-72B & 2024.43 & Qwen & pre & no \\
Claude-3.5-Sonnet-20240620 & 2024.47 & Anthropic & pre & no \\
Mistral-Large & 2024.55 & Mistral & pre & no \\
GPT-4O-mini-2024-07-18 & 2024.55 & OpenAI & pre & no \\
GPT-4O-2024-08-06 & 2024.60 & OpenAI & pre & no \\
O1-Preview-2024-09-12 & 2024.70 & OpenAI & post & yes \\
O1-Mini-2024-09-12 & 2024.70 & OpenAI & post & yes \\
Qwen2.5-Ins-32B & 2024.72 & Qwen & post & no \\
Qwen2.5-Ins-7B & 2024.72 & Qwen & post & no \\
Qwen2.5-Ins-72B & 2024.72 & Qwen & post & no \\
Gemini-Flash-1.5-002 & 2024.73 & Google & post & no \\
Gemini-Pro-1.5-002 & 2024.73 & Google & post & no \\
Claude-3.5-Sonnet-20241022 & 2024.81 & Anthropic & post & no \\
DeepSeek-R1-Lite-Preview & 2024.87 & DeepSeek & post & yes \\
Qwen2.5-Coder-Ins-7B & 2024.87 & Qwen & post & no \\
Qwen2.5-Coder-Ins-32B & 2024.87 & Qwen & post & no \\
QwQ-32B-Preview & 2024.89 & Qwen & post & yes \\
Gemini-Exp-1206 & 2024.93 & Google & post & no \\
Gemini-Flash-2.0-Thinking-12-19 & 2024.96 & Google & post & yes \\
LLama3.3-70b-Ins & 2024.96 & Meta & post & no \\
Gemini-Flash-2.0-Thinking & 2024.96 & Google & post & yes \\
O1-2024-12-17 (High) & 2024.96 & OpenAI & post & yes \\
O1-2024-12-17 (Low) & 2024.96 & OpenAI & post & yes \\
Gemini-Flash-2.0-Exp & 2024.96 & Google & post & no \\
O1-2024-12-17 (Med) & 2024.96 & OpenAI & post & yes \\
DeepSeek-V3 & 2024.98 & DeepSeek & post & no \\
DeepSeek-R1-Preview & 2025.05 & DeepSeek & post & yes \\
Gemini-Flash-2.0-Thinking-01-21 & 2025.06 & Google & post & yes \\
O3-Mini-2025-01-31 (High) & 2025.08 & OpenAI & post & yes \\
O3-Mini-2025-01-31 (Low) & 2025.08 & OpenAI & post & yes \\
O3-Mini-2025-01-31 (Med) & 2025.08 & OpenAI & post & yes \\
QwQ-Max-Preview & 2025.13 & Qwen & post & yes \\
Kimi-k1.6-IOI-high & 2025.15 & Moonshot & post & yes \\
Kimi-k1.6-IOI & 2025.15 & Moonshot & post & yes \\
Claude-3.7-Sonnet & 2025.15 & Anthropic & post & no \\
Llama-3\_1-Nemotron-Nano-8B-v1 & 2025.20 & NVIDIA & post & yes \\
Gemini-2.5-Pro-03-25 & 2025.23 & Google & post & no \\
QwQ-32B\_temp & 2025.23 & Qwen & post & yes \\
DeepCoder-14B-Preview & 2025.27 & DeepSeek & post & yes \\
Llama-3\_1-Nemotron-Ultra-253B-v1 & 2025.27 & NVIDIA & post & yes \\
O4-Mini (Medium) & 2025.29 & OpenAI & post & yes \\
O4-Mini (High) & 2025.29 & OpenAI & post & yes \\
O4-Mini (Low) & 2025.29 & OpenAI & post & yes \\
O3 (High) & 2025.29 & OpenAI & post & yes \\
Gemini-2.5-Flash-04-17 & 2025.29 & Google & post & no \\
Qwen3-235B-A22B & 2025.32 & Qwen & post & yes \\
Gemini-2.5-Pro-05-06 & 2025.35 & Google & post & no \\
Gemini-2.5-Flash-05-20 & 2025.38 & Google & post & no \\
Claude-Sonnet-4 & 2025.39 & Anthropic & post & no \\
Claude-Opus-4 & 2025.39 & Anthropic & post & no \\
Claude-Opus-4 (Thinking) & 2025.39 & Anthropic & post & yes \\
Claude-Sonnet-4 (Thinking) & 2025.39 & Anthropic & post & yes \\
DeepSeek-R1-0528 & 2025.40 & DeepSeek & post & yes \\
Gemini-2.5-Pro-06-05 & 2025.43 & Google & post & no \\
Grok-3-Mini (High) & 2025.44 & xAI & post & yes \\
EXAONE-4.0-32B & 2025.54 & LG & post & yes \\
OpenReasoning-Nemotron-32B & 2025.54 & NVIDIA & post & yes \\
\bottomrule
\end{longtable}
}

\end{document}